\documentclass[10pt,twocolumn,letterpaper]{article}

\usepackage[pagenumbers]{cvpr} % To force page numbers, e.g. for an arXiv version

\definecolor{cvprblue}{rgb}{0.21,0.49,0.74}
\usepackage[pagebackref,breaklinks,colorlinks,allcolors=cvprblue]{hyperref}

\usepackage{booktabs}
\usepackage{multirow}
\usepackage{makecell}
\usepackage{siunitx}
\usepackage[table]{xcolor} % 若不需要着色，可删去并删除 \rowcolor/\cellcolor

\usepackage{tikz}

\title{Denoising Vision Transformer Autoencoder with Spectral Self-Regularization}

\author{
Xunzhi Xiang$^{1}\thanks{This work was conducted during the author's intership at Kling Team, Kuaishou Technology}\,\,\footnotemark[2]$, Xingye Tian$^{2}\thanks{Equal Contribution}$, Guiyu Zhang$^{3}$, Yabo Chen$^{4}$, Shaofeng Zhang$^{5}$ \\
Xuebo Wang$^{2}$, Xin Tao$^{2}$, Qi Fan$^{1}\thanks{Corresponding author}$ \\
$^{1}$Nanjing University, $^{2}$Kling Team, Kuaishou Technology \\
$^{3}$Chinese University of Hong Kong, Shenzhen, $^{4}$Shanghai Jiao Tong University \\
$^{5}$University of Science and Technology of China
}

\begin{document}
\maketitle
\begin{abstract}
Variational autoencoders (VAEs) typically encode images into a compact latent space, reducing computational cost but introducing an optimization dilemma: a higher-dimensional latent space improves reconstruction fidelity but often hampers generative performance. Recent methods attempt to address this dilemma by regularizing high-dimensional latent spaces using external vision foundation models (VFMs). However, it remains unclear how high-dimensional VAE latents affect the optimization of generative models. To our knowledge, our analysis is the first to reveal that redundant high-frequency components in high-dimensional latent spaces hinder the training convergence of diffusion models and, consequently, degrade generation quality. To alleviate this problem, we propose a spectral self-regularization strategy to suppress redundant high-frequency noise while simultaneously preserving reconstruction quality. The resulting Denoising-VAE, a ViT-based autoencoder that does not rely on VFMs, produces cleaner, lower-noise latents, leading to improved generative quality and faster optimization convergence. We further introduce a spectral alignment strategy to facilitate the optimization of Denoising-VAE-based generative models. Our complete method enables diffusion models to converge approximately 2$\times$ faster than with SD-VAE, while achieving state-of-the-art reconstruction quality (rFID = 0.28, PSNR = 27.26) and competitive generation performance (gFID = 1.82) on the ImageNet 256$\times$256 benchmark.

\end{abstract}

\section{Introduction}
\label{sec:intro}

\begin{figure*}[t]
  \hfill
  % \fbox{\rule{0pt}{2in} \rule{\linewidth}{0pt}}
  \includegraphics[width=\linewidth]{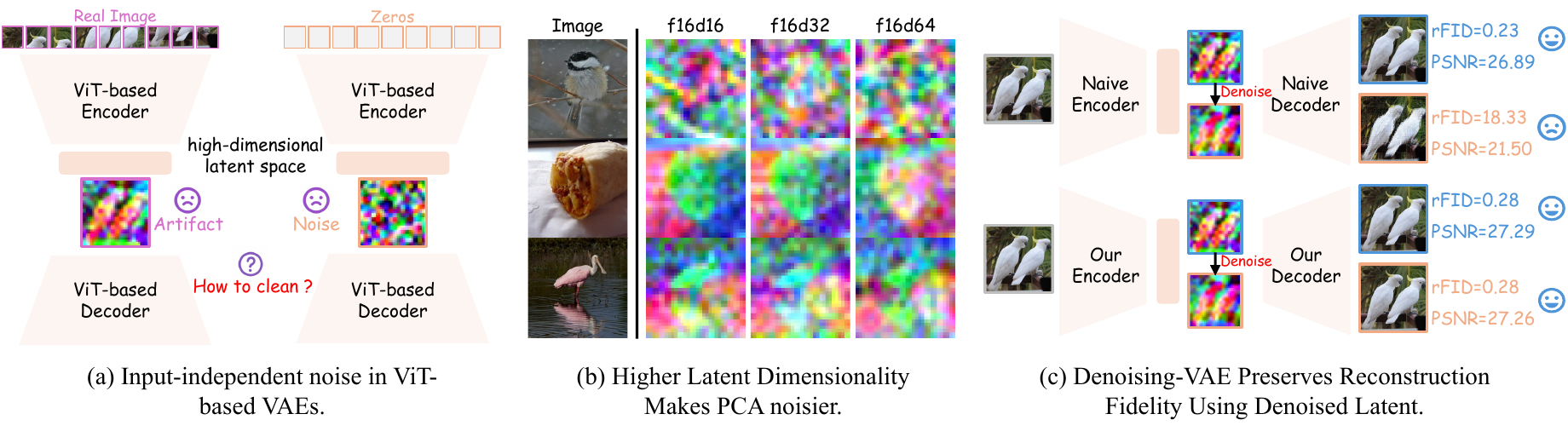}
  \vspace{-2em}
  \caption{Frequency-domain analysis of ViT-based VAE latent spaces and comparison with convolutional baselines.
    (a) Encoding a uniform-color image with a ViT-based tokenizer yields structured high-frequency artifacts in the latent space, revealing input-independent noise injected during tokenization.
    (b) As latent dimensionality increases, PCA projections of ViT-based latents become increasingly noisy and spatially unstable, indicating the amplification of undesired high-frequency variation.
    (c) In contrast, Conventional VAEs show strong dependence on high-frequency latent signals, where reconstruction fidelity is tightly coupled to noise patterns.}
  \label{fig:teaser}
  \vspace{-1em}
\end{figure*}

% 1. 指出问题
Latent diffusion models \cite{LDN, SDXL, FastDIT, FLUX} %and autoregressive models \cite{tamingTrans, llamagen} 
have demonstrated strong performance in high-quality image and video synthesis.
These models typically operate in compressed latent spaces to reduce computational costs associated with high-resolution image generation.

Diffusion models generally comprise two sequential modules.
First, a variational autoencoder (VAE) \cite{VAE}, also referred to as a visual tokenizer \cite{AE, mae}, encodes the input image into a compact, low-resolution latent representation.
Then, a generative model \cite{maskgit, dit, sit} is trained on the encoded latent space \cite{diffvae, imagefolder}, reducing computational overhead and enabling high-quality image generation.
However, VAEs face a trade-off dilemma \cite{dilemma1, dilemma2, vavae, DCAE15}: higher-dimensional latent representations improve reconstruction quality for diffusion models,
but the resulting high computational complexity usually impedes optimization and consequently degrades generative performance.

% 2. 现在的方法怎么解决问题，以及不足
% Some works \cite{vavae, maetok} attempt to mitigate this problem by employing vision foundation models \cite{mae, dino, IBOT, CLIP, dinov2} to regularize high-dimensional latent spaces. 
Some works \cite{vavae, maetok} attempt to mitigate this problem by employing VFMs \cite{mae, dino, IBOT, CLIP, dinov2} to regularize high-dimensional latent spaces. 
Despite empirical success, the root cause of this trade-off remains unclear, limiting a deeper understanding of diffusion models.
%poorly understood and lacks clear interpretability.

% 3. 问题背后的原因是什么
% First, we encode a uniform-color image using the VAE encoder and visualize its latent representation via principal component analysis (PCA).
% Ideally, the PCA result should be smooth and structureless \cite{DVT}. 
% However, as shown in Figure~\ref{fig:teaser}, the result exhibits prominent artifacts, indicating that the encoding process injects input- independent high-frequency noise.
% Second, we further encode the same image using VAEs of different dimensionalities.
% The PCA results of higher-dimensional VAEs exhibit stronger artifacts, suggesting that increased latent dimensionality leads to stronger high-frequency noise.
To investigate this issue, we conduct a preliminary analysis.
We use a VAE to encode a uniform-color image and visualize its latent representation using principal component analysis (PCA).
As shown in Figure~\ref{fig:teaser}, the encoded latent representation exhibits noticeable artifacts.
As latent dimensionality increases, the artifacts become more pronounced, suggesting that the encoder introduces input-independent high-frequency noise—an effect further amplified in higher-dimensional latent spaces.

%Ideally, the PCA result should be smooth and structureless \cite{DVT}, but Figure~\ref{fig:teaser} shows clear artifacts, indicating that the encoder introduces input‑independent high‑frequency noise.
% Then, we further repeat the experiment on the same image using VAEs with different latent dimensionalities, and observe that higher-dimensional latents exhibit increasingly pronounced artifacts, suggesting that larger latent spaces amplify high-frequency noise.
%These observations indicate that: \textit{The latent space of ViT-based VAE contains persistent high-frequency noise that intensifies with dimensionality.}

% 3. 问题背后的原因是什么
We hypothesize that high-frequency noise increases optimization complexity, thereby hindering the convergence of diffusion models.
To address this, we propose actively denoising and smoothing the latent space while preserving reconstruction quality, yielding well-structured high-dimensional VAE latents for high-quality image generation.
Specifically, we introduce Denoising-VAE, a ViT-based \cite{transformer, ViT}, VFM-free autoencoder that regularizes the high-dimensional latent space through Spectral Self-Regularization.

% 4. 我们与其他方法不同和优势，我们怎么解决问题
%Motivated by this insight, we propose Denoising-VAE, a ViT-based \cite{transformer, ViT}, VFM-free autoencoder that regularizes the high-dimensional latent space via Spectral Self-Regularization. 
% 5. 解决问题的具体做法
In Denoising-VAE, spectral regularization is applied to the latent space during autoencoder training.
Spectral regularization encourages the model to reconstruct high-quality images from both original and low-pass filtered latent representations.
%enabling the model to reconstruct high-quality images from the original latent representations, and simultaneously produce a perceptually equivalent image from a low-pass filtered version of the same latent. 
This constraint suppresses redundant high-frequency noise while maintaining reconstruction quality in VAEs.
It enables VAEs to produce cleaner, lower-noise high-dimensional latents, leading to improved generative performance and faster optimization convergence.
In addition, we propose Frequency-Aware Diffusion Alignment (FDA) to accelerate the optimization of Denoising-VAE-based diffusion models.
During generative model training, FDA generates low-pass filtered latent representations at multiple frequency bands and aligns noisy latents with their corresponding clean counterparts.

%enabling the model to learn the interdependencies between diverse scale features. 
%leverages the leverage the cleaner latents to guide the
%is a coarse-to-fine alignment scheme: a clean latent obtained under strong low-pass filtering serves as a reference to guide the alignment of noisier latents obtained under weaker filtering. 
%This progressive supervision  accelerates the training in high-dimensional spaces.
%Finally, we validate the effectiveness of our approach on the ImageNet dataset. Experimental results demonstrate that our method achieves competitive performance in both reconstruction and generation.

Unlike prior VAE approaches~\cite{vavae, maetok}, we identify high-frequency noise as the root cause of the optimization dilemma in high-dimensional latent spaces and mitigate it through lightweight spectral self-regularization.
On the ImageNet 256×256 benchmark, Denoising-VAE achieves state-of-the-art reconstruction performance (rFID: 0.28, PSNR: 27.26), outperforming the best ViT-based alternative~\cite{Textok} by 2.88 dB in PSNR, while maintaining competitive generative performance (gFID: 1.82).
Additionally, our high-dimensional VAE (32 channels) enables diffusion models to converge nearly $2\times$ faster than the conventional convolutional SD-VAE~\cite{LDN} (4 channels), while reducing total autoencoder GFLOPs by $5.75\times$. 
These results demonstrate the efficiency and scalability advantages of our spectral regularization method.

Overall, our work delivers the following contributions:

\begin{itemize}
  \item We empirically show that ViT-based VAEs exhibit persistent high-frequency noise in the latent space, which intensifies with increasing latent dimensionality.
  \item We introduce Spectral Regularization to denoise the latent space, yielding cleaner, structurally constrained latents and easing diffusion optimization.
  \item We propose Frequency-Domain Diffusion Alignment accelerating diffusion convergence.
\end{itemize}

\section{Related Work}
\label{sec:Related}

\noindent \textbf{Latent Generative Models.~}
Latent generative models have established a foundation for high-fidelity image synthesis. Continuous VAEs \cite{VAE, robustvae, vae2} learn smooth latent spaces with Gaussian priors, while VQ-VAE models \cite{VQVAE, softvq} discrete representations using a vector-quantized codebook coupled with an autoregressive decoder. Following the discrete paradigm, VQGAN~\cite{tamingTrans} enhances reconstruction through adversarial training and transformer-based autoregressive modeling. MaskGIT \cite{maskgit} further accelerates the generation of discrete latents by introducing a masked token prediction schedule that enables parallel decoding. Alternatively, Latent Diffusion Models (LDM) \cite{LDN, SDXL} apply diffusion processes directly in a compressed latent space, enabling efficient and scalable high-resolution synthesis. Among these, the Diffusion Transformer (DiT) \cite{dit} demonstrates the strong suitability of transformer architectures for diffusion-based generation. Building on DiT, the SiT \cite{sit} model incorporates flexible interpolation mechanisms for versatile distribution mapping and improved modularity.

\noindent \textbf{Image Tokenizer.~}
Image tokenizers map images into latents that are simpler to handle than pixels in generative modeling. Early image tokenizers can be grouped into continuous and discrete two classes. Continuous tokenizers, typically variational autoencoders, learn Gaussian-prior latent spaces at reduced spatial resolution and are widely used for latent diffusion. In contrast, discrete tokenizers \cite{GigaTok, AliTok} such as VQ-VAE \cite{VQVAE} instantiate vector-quantized codebooks and emit index sequences for autoregressive decoders. Recent work targets higher compression ratios and improved reconstruction fidelity while preserving downstream generative performance. DC-AE \cite{DCAE} and DC-AE-1.5 \cite{DCAE15} increase the spatial compression ratio to improve diffusion efficiency. VAVAE \cite{vavae} enlarges the latent channel dimensionality and uses vision-foundation-model regularization to enhance reconstruction, while EQ-VAE \cite{EQVAE} applies scale equivariance regularization to improve compatibility with diffusion training. In contrast, we analyze the reconstruction–generation trade-off from a frequency viewpoint and find that the degradation at higher dimensionality is driven by excess high-frequency energy. We therefore introduce Spectral Regularization to denoise the latent space, suppressing irrelevant high-frequency components and improving diffusion optimization without external VFMs.

% \noindent \textbf{Latent Equivariance.~}
% To reduce latent variation and improve compatibility with generative models, recent works commonly enforce \emph{latent-space equivariance} to predefined \emph{spatial transformations}. These methods \cite{EQVAE, improve} explicitly encourage tokenizers to produce representations that transform in a predictable and structured way under geometric changes, thereby yielding more generator-friendly latents. However, such approaches primarily focus on geometric consistency and overlook the inherent noise and instability within high-dimensional latent spaces, which can still hinder generative optimization.

% \noindent \textbf{Our Method.~} 
% Consider an image tokenizer with encoder $\mathcal{E}:\mathcal{X}\to\mathcal{Z}$ and decoder $\mathcal{D}:\mathcal{Z}\to\mathcal{X}$. 
% Our goal is not to reuse transformed latents, but to actively denoise the latent representation $\mathcal{E}(x)$ itself.
% We introduce frequency-domain suppression to remove redundant high-frequency components while ensuring that the denoised latent can still reconstruct a perceptually equivalent image.
% This design fundamentally differs from prior equivariance-based approaches: rather than enforcing structural invariance, we aim to obtain cleaner and more stable latents that facilitate downstream generative optimization without compromising reconstruction quality.

\noindent \textbf{Latent Equivariance.~}
A closely related line of work \cite{EQVAE, improve} enforces \emph{spatial equivariance} in the latent space to obtain generator-friendly representations through predictable geometric transformations. 
In contrast, our objective is to directly denoise the latent representation by suppressing redundant high-frequency noise while ensuring that the denoised latent can still reconstruct a perceptually equivalent image. 
This frequency-domain formulation focuses on improving latent quality and stability, providing cleaner representations that better support generative modeling without compromising reconstruction fidelity.

% \noindent \textbf{Our Method.~}
% Instead of constraining geometric behavior, we approach the problem from a frequency perspective. 
% Our goal is to directly denoise the latent representation $\mathcal{E}(x)$ by suppressing redundant high-frequency components that hinder optimization, while ensuring that the cleaned latent can still reconstruct a perceptually equivalent image. 
% This frequency-domain formulation differs conceptually and practically from equivariance-based approaches. 
% Rather than enforcing spatial consistency, we explicitly enhance latent quality, producing representations that are cleaner, more stable, and better aligned with generative training without compromising reconstruction fidelity.

\section{Denoising-VAE}
\label{sec:Method}

\subsection{Multi-Level Spectral Regularization}
\label{sec:mlsr}
% This section aims to operate directly in the latent space by applying low-pass filtering to suppress high-frequency noise and obtain cleaner representations. The denoised latents can still reconstruct images. However, simultaneously requiring both the original and denoised latents to reconstruct the original image often leads to unstable training for VAEs. Inspired by human visual perception, which is more sensitive to low-frequency distortions than to high-frequency details, we relax the reconstruction requirement of denoised latents.
% They only need to reconstruct perceptually equivalent images instead of exactly matching the original input.

This section aims to directly simplify the high-dimensional latent space by applying low-pass filtering to suppress high-frequency noise and obtain cleaner representations. The denoised latents are then used for image reconstruction. However, in conventional VAE training, requiring both the original and denoised latents to reconstruct the exact same input image often leads to unstable optimization. Motivated by human visual perception, we relax this constraint. Human vision is more sensitive to low-frequency distortions, such as errors in object contours or shading, than to high-frequency details like fine textures. Therefore, we only require the denoised latents to reconstruct images that are perceptually equivalent to the original, rather than exactly matching it pixel by pixel.

% To realize this perceptual reconstruction strategy, we propose \emph{Multi-Level Spectral Regularization (MLSR)} to explicitly suppress high-frequency energy in the latent space. Unlike previous approaches that concentrate solely on spatial equivariance, MLSR extends the constraint to the spectral domain, encouraging latent representations to remain perceptually consistent under frequency-specific transformations. Specifically, MLSR enforces spectral consistency across multiple levels by aligning latents that undergo controlled high-frequency attenuation with image reconstructions filtered to the corresponding low-frequency bands. This approach guides the latent space toward smoother and more structured representations that retain perceptually important information while discarding redundant high-frequency detail, thereby improving both reconstruction and generation performance.
To realize this perceptual reconstruction strategy, we propose \emph{Multi-Level Spectral Regularization (MLSR)}. 
MLSR operates in the frequency domain to progressively suppress redundant high-frequency noise in the latent space while preserving perceptually important low-frequency information. 
At each level, it aligns denoised latents with their corresponding low-frequency reconstructions, encouraging spectral consistency and stabilizing optimization.  This process suppresses redundant high-frequency noise that hinders convergence, resulting in cleaner latents and improved generative performance.

\noindent\textbf{Multi-level Spectral Regularization.}
For each training image \(x\), we define a pair of \emph{matched} operators indexed by an attenuation level \(\ell\in\{0,1,2,3\}\):
a latent-space operator \(\mathcal{S}_\ell\) that suppresses high-frequency energy, and an image-space operator \(\mathcal{G}_\ell\) that applies a perceptually calibrated low-pass of the equivalent strength. Specifically, let $z = \mathcal{E}(x)$ denote the latent representation of the input $x$, and let $\mathcal{D}(z)$ denote the reconstruction of input image.
At each training step, we \emph{uniformly} sample one level,
\begin{equation}
\ell \sim \mathrm{Unif}\{0,1,2,3\},
\end{equation}
and ensure the decoded image from the attenuated latent matches the spectrally simplified original:
\begin{equation}
\mathcal{L}_{\text{spec}}
\;=\;
d\!\left(
(\mathcal{D}(\mathcal{S}_\ell(z))\big),\;
\mathcal{G}_\ell(x)
\right),
\end{equation}
where \(d(\cdot,\cdot)\) is a weighted combination of pixel-wise MSE loss, perceptual loss \cite{Perceptual-Losses, VGG, convnet}, adversarial (GAN) loss \cite{tamingTrans}, and a latent-space KL loss.
% This stochastic, single-level objective pairs latent high-frequency attenuation with a matched image-space target, denoising the latent representation while preserving perceptually salient structure.

\noindent\textbf{Latent operator.}  
We suppress high-frequency components in the latent space while preserving low frequencies.  
Given a latent representation \(z\), we:  
(i) apply a 2D Fourier transform $\mathcal{F}$,  
(ii) attenuate only the highest-magnitude frequency components according to the attenuation level,  
and (iii) apply the inverse transform $\mathcal{F}^{-1}$ to map back to the spatial domain.
This is implemented as:
\begin{equation}
\mathcal{S}_\ell(z) = \mathcal{F}^{-1}\big( \mathcal{M}_\ell \odot \mathcal{F}(z) \big),
\end{equation}
where \(\odot\) denotes element-wise multiplication, and \(\mathcal{M}_\ell\) is a frequency-domain mask that preserves all low-frequency components.  In practice, we choose \(\ell\) uniformly from \(\{0, 1, 2, 3\}\), corresponding to 0\%, 25\%, 50\%, and 75\% high-frequency attenuation.  
This operation reduces latent noise while retaining perceptually important structure.

\noindent\textbf{Image operator.}
To ensure perceptual consistency between latent-space suppression and the image domain, we create a corresponding reference image using adaptive Gaussian blurring. This is implemented as:
\begin{equation}
  \mathcal{G}_\ell(x) \;=\; \operatorname{GaussianBlur}(x;\,\sigma_\ell).
\end{equation}
Implementation-wise, we blur each channel using a separable Gaussian with reflection padding to avoid edge artifacts. 
We choose $\ell$ uniformly from $\{0,1,2,3\}$, corresponding to blur levels $\sigma_\ell = 0,\; 0.05,\; 0.25,\; 0.5$. 
This ensures the image target aligns with the degree of latent denoising and discourages the decoder from overfitting to imperceptible details.

\subsection{Denoising Latents for Generation}
\label{sec:denoise}
% While existing equivariance-based regularization methods improve the structural coherence of the latent space, they suffer from a critical limitation: the transformed latent representations obtained through equivariant transformations are  discarded during training. As a result, generative models are still supervised on the original, unprocessed latent codes, failing to fully leverage the structural benefits inherent in the equivariant framework. Through Multi-Level Spectral Regularization proposed in Sec.~\ref{sec:mlsr}, we establish a rigorous frequency-domain equivariance between the latent and image spaces. This property enables the diffusion models use of high-frequency-attenuated latent representations. By suppressing high-frequency noise, these modified latents provide a cleaner and more learnable representation for generative modeling. We therefore propose to shift the training objective of generative models from the original latent space to this spectrally purified latent domain.

Latent representations in high-dimensional VAEs contain large amounts of redundant high-frequency noise. This noise does not contribute to image semantics but significantly complicates the learning dynamics of diffusion models, which must implicitly fit both meaningful structure and input-independent artifacts. Approaches grounded in equivariance focus primarily on geometric consistency and do not address the fundamental issue of noisy latent distributions, since the original latents remain the supervision target throughout training.
Our approach takes a different perspective: instead of regularizing latent geometry, we directly clean the latent space. Multi-Level Spectral Regularization (Sec.~\ref{sec:mlsr}) removes high-frequency components in a controlled, progressive manner, yielding latent codes that are smoother, more structured, and substantially easier to model. By training diffusion models on these spectrally denoised latents, we reduce optimization complexity and improve both stability and generation quality.

Specifically, we obtain a denoised latent representation $z^{(\ell)}=\mathcal{S}_\ell(z)$ from the original $z=\mathcal{E}(x)$ by suppressing its high-frequency components. The diffusion model is trained directly on this denoised latent rather than the raw, noise-contaminated one.  Concretely, in the noise-prediction loss, we replace $z$ with $z^{(\ell)}$ at each training step. This process can be expressed as:
\begin{equation}
\label{eq:denoise-latent-diff}
\min_{\theta}\;\mathbb{E}_{\ell,\,t,\,\epsilon}\Big[
\big\|\epsilon_{\theta}\!\big(\sqrt{\bar{\alpha}_{t}}\,z^{(\ell)}+\sqrt{1-\bar{\alpha}_{t}}\,\epsilon,\,t,\,c\big)-\epsilon\big\|_{2}^{2}
\Big].
\end{equation}
where $\epsilon_{\theta}(\cdot)$ is the network for predicting noise. $\epsilon$ represents the actual noise. $c$ denotes the conditional information. $t$ represents the timestep of the denoising process.

%====================== Two-column table (right-aligned numbers; no citations) =======================
\begin{table*}[t]
\centering
% \small
\footnotesize
\setlength{\tabcolsep}{5.5pt}
\renewcommand{\arraystretch}{1.12}
\caption{
Reconstruction and generation performance. 
% Reconstruction metrics are computed from autoencoder outputs; generation without classifier-free guidance (CFG) \cite{CFG} reports gFID$\downarrow$, sFID$\downarrow$, IS$\uparrow$, Precision (Pre.$\uparrow$), and Recall (Rec.$\uparrow$). 
Notably, Denoising-VAE significantly improves generation quality while preserving reconstruction fidelity, with both 16- and 32-channel settings outperforming the original SD-VAE baseline. \noindent$^{\dagger}$ Indicates using their VAE checkpoints, while the SiT training setting is kept identical to ours for a fair comparison. $^{\ddagger}$ Indicates results reported in the original paper.
}
\begin{tabular}{lccccccccccc}
\toprule
\multirow{2}{*}{Tokenizer} &
\multicolumn{1}{c}{Speed} &
\multicolumn{4}{c}{Reconstruction Performance} &
\multicolumn{5}{c}{Generation Performance} \\
\cmidrule(lr){2-2} \cmidrule(lr){3-6} \cmidrule(lr){7-11}
& \textbf{GFLOPs}$\downarrow$
& \textbf{rFID}$\downarrow$ & \textbf{PSNR}$\uparrow$ & IS$\uparrow$ & SSIM$\uparrow$
& \textbf{gFID}$\downarrow$ & sFID$\downarrow$ & IS$\uparrow$ & Pre.$\uparrow$ & Rec.$\uparrow$ \\
\midrule
\rowcolor{red!6}\multicolumn{11}{l}{\textit{\textbf{Conv-based tokenizer - f8d4}}} \\
\midrule
SD-VAE$^{\dagger}$~\cite{LDN}                                        & 445  & 0.62 & 26.04 & --     & 0.834 & 18.01 & 5.13 & 74.29 & 0.64 & 0.64 \\
EQ-VAE$^{\ddagger}$~\cite{EQVAE}                                      & 445  & 0.82 & 25.95 & --     & 0.720 & 16.10 & -- & -- & -- & -- \\
\midrule
\rowcolor{red!6}\multicolumn{11}{l}{\textit{\textbf{ViT-based tokenizer - f16d16}}} \\
\midrule
MLSR                                       & 87.61  & 0.62 & 25.11 & 217.39 & 0.737    & 16.98 & 5.12 & 74.20 & 0.66 & 0.62 \\
MLSR + Denoising                           & 87.61  & 0.63 & 25.12 & 217.02 & 0.736    & 14.00 & 4.96 & 84.14 & 0.68 & 0.61 \\
MLSR + Denoising + FDA                     & 87.61  & 0.63 & 25.12 & 217.02 & 0.736    & 11.58 & 4.74 & 93.18 & 0.70 & 0.62 \\
\midrule
\rowcolor{red!6}\multicolumn{11}{l}{\textit{\textbf{ViT-based tokenizer - f16d32}}} \\
\midrule
MLSR                                       & 87.63  & 0.28 & 27.29 & 233.42 & 0.815    & 19.26 & 5.48 & 68.43 & 0.63 & 0.63 \\
MLSR + Denoising                           & 87.63  & 0.28 & 27.26 & 233.47 & 0.814    & 17.65 & 5.35 & 73.26 & 0.64 & 0.63 \\
MLSR + Denoising + FDA                     & 87.63  & 0.28 & 27.26 & 233.47 & 0.814    & 14.34 & 5.09 & 83.28 & 0.67 & 0.62 \\
\midrule
\rowcolor{red!6}\multicolumn{11}{l}{\textit{\textbf{ViT-based tokenizer - f16d64}}} \\
\midrule
MLSR                                       & 87.66  & 0.19 & 29.45 & 239.26 & 0.878 & 24.22 & 6.22 & 57.07 & 0.60 & 0.62 \\
MLSR + Denoising                           & 87.66  & 0.19 & 29.40 & 239.28 & 0.878 & 23.72 & 6.16 & 58.63 & 0.60 & 0.63 \\
MLSR + Denoising + FDA                     & 87.66  & 0.19 & 29.40 & 239.28 & 0.878 & 20.87 & 5.83 & 64.09 & 0.62 & 0.64 \\
\bottomrule
\end{tabular}
\label{tab:regularization}
\end{table*}
%====================== end table =======================

\begin{figure}[t]
  \hfill
  % \fbox{\rule{0pt}{2in} \rule{\linewidth}{0pt}}
  \includegraphics[width=\linewidth]{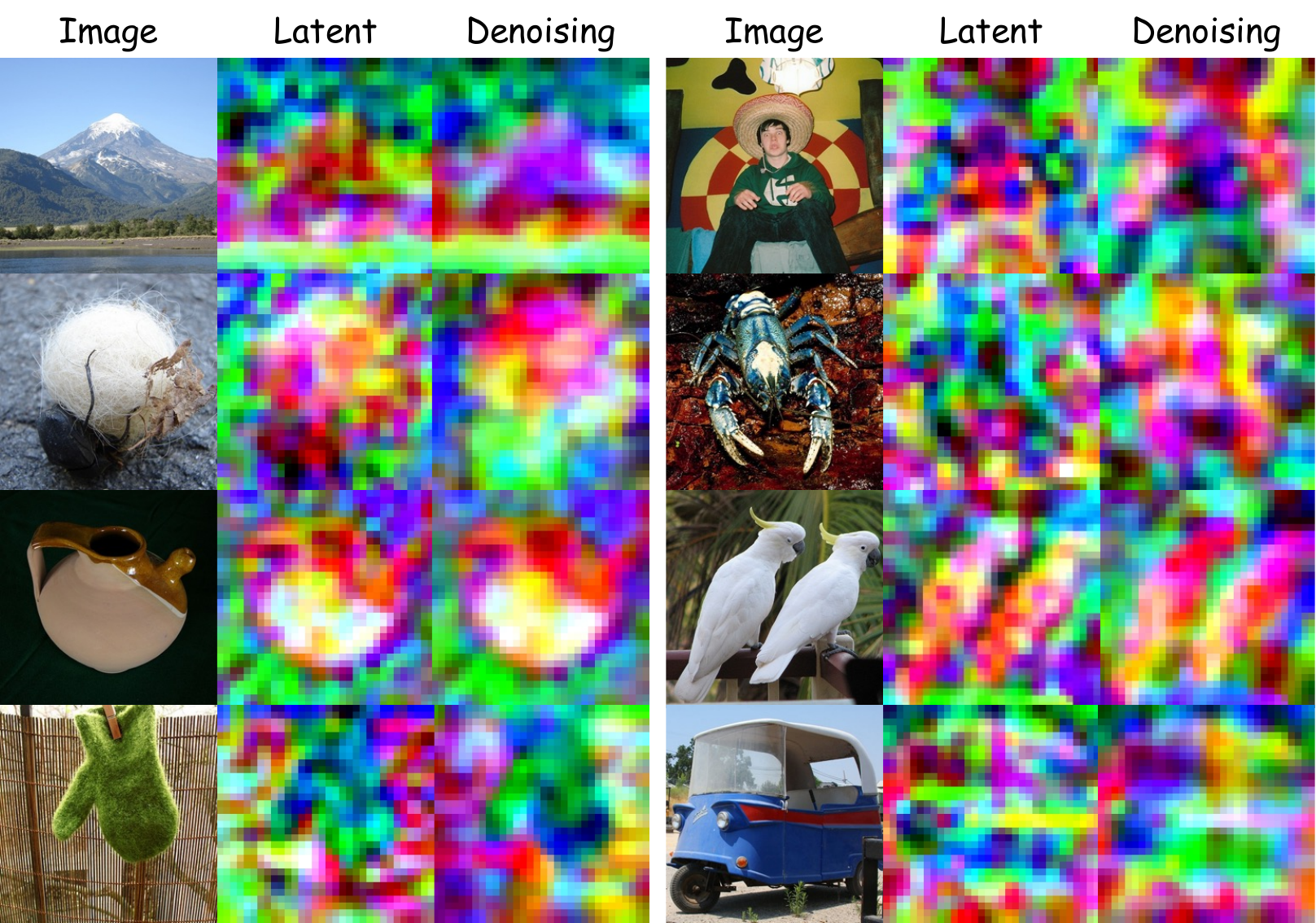}
  \caption{Per-image latent denoising with Denoising-VAE. After Spectral Regularization reveal smoother, more coherent structure.}
  \label{fig:pca-visualization}
  \vspace{-1.3em}
\end{figure}

\subsection{Frequency-Domain Diffusion Alignment}
\label{sec:fda}
Given a multi-level latent space, we accelerate training by letting a \emph{cleaner, spectrally simpler} latent guide a \emph{noisier, more complex} latent. 
Let \(z^{(\ell)}=\mathcal{S}_\ell(\mathcal{E}(x))\) be the level-\(\ell\) attenuated latent, where larger \(\ell\) suppresses more high-frequency energy. 
At each iteration we sample \((\ell_{\text{hi}},\ell_{\text{lo}})\) with \(\ell_{\text{hi}}>\ell_{\text{lo}}\), a diffusion timestep \(t\), and noise \(\epsilon\sim\mathcal{N}(0,I)\). 
The student sees a noisy, lower-attenuation latent,
\begin{equation}
\tilde z_t \;=\; \sqrt{\bar\alpha_t}\,z^{(\ell_{\text{lo}})} \;+\; \sqrt{1-\bar\alpha_t}\,\epsilon,
\end{equation}
while the teacher sees a clean, higher-attenuation latent \(\hat z = z^{(\ell_{\text{hi}})}\) at \(t{=}0\). 
Intuitively, the teacher provides a spectrally simplified target that emphasizes global structure, giving the student a coarse-to-fine curriculum consistent with diffusion’s noise schedule.
Let \(\epsilon_\theta(\cdot,t,c)\) be the diffusion network and \(\phi(\cdot,t,c)\) a selected intermediate feature. 
We follow SRA \cite{SRA} to instantiate the teacher as an EMA of the student parameters, denoted \(\theta^{-}\). 
The alignment loss matches student features to teacher features:
\begin{equation}
\label{eq:fda_align}
\mathcal{L}_{\text{align}}
\;=\;
\big\|\;\phi_{\theta}(\tilde z_t,\,t,\,c)\;-\;\phi_{\theta^{-}}(\hat z,\,0,\,c)\;\big\|_2^2,
\end{equation}
where the teacher branch is stop-gradient by construction.
We add this alignment to the standard diffusion objective:
\begin{equation}
\label{eq:fda_total}
\mathcal{L}\;=\;\mathcal{L}_{\text{diff}} \;+\; 
\lambda_{\text{align}}\,\mathcal{L}_{\text{align}}.
\end{equation}
We place higher weight on early timesteps via \(\lambda_{\text{align}}(t)\) to respect the coarse-to-fine schedule. 
The design accelerate the optimization of Denoising-VAE-based diffusion models.

\section{Experiment}
% In this section, we systematically investigate how denoising latent spaces of varying dimensionalities affects both reconstruction and generation performance, aiming to better understand and effectively address the underlying optimization challenges.

% \subsection{settings}
\noindent\textbf{Training.}
% We conduct all experiments on ImageNet \cite{Imagenet}, which contains 1.28 million training images and 50 thousand validation images. For the reconstruction stage, the tokenizer adopts a ViT encoder–decoder backbone and is trained with a learning rate of \(1\times 10^{-4}\) under a global batch size of 512. We evaluate three configurations, \textbf{\textit{f16d16}}, \textbf{\textit{f16d32}}, and \textbf{\textit{f16d64}}, where \(f\) denotes the spatial downsampling factor and \(d\) denotes the number of latent channels. Under this setup, the Denoising\mbox{-}VAE is trained for 500K steps. For the generation stage, we use SiT\mbox{-}XL \cite{sit} as the diffusion model and adopt the same tokenizer configurations. To reduce memory consumption and increase training efficiency, we precompute latent representations using the frozen Denoising\mbox{-}VAE and train SiT\mbox{-}XL directly on these latents without data augmentation. SiT\mbox{-}XL is trained for 800 epochs in the main experiments and 80 epochs in ablation studies. Unless otherwise noted, all other hyperparameters are held constant across settings for fair comparison.
All experiments are conducted on the ImageNet dataset \cite{Imagenet}, which contains 1.28M training images and 50K validation images.
In the reconstruction stage, the tokenizer adopts a ViT-based encoder-decoder architecture and is trained with a batch size of 512 and a learning rate of $1\times10^{-4}$. 
We evaluate three configurations: {\textit{f16d16}}, {\textit{f16d32}}, and {\textit{f16d64}}, where $f$ denotes the spatial downsampling factor and $d$ denotes the number of latent channels.
Under these settings, the Denoising-VAE is trained for 500K steps.
In the generation stage, we use SiT-XL \cite{sit} as the diffusion model with the same tokenizer configurations.
To reduce memory usage and improve training throughput, we precompute latent representations using the frozen Denoising-VAE and train SiT-XL directly on these latents without data augmentation.
SiT-XL is trained with a batch size of 256 and a learning rate of $1\times10^{-4}$.
SiT\mbox{-}XL is trained for 800 epochs in the main experiments and 80 epochs in ablation studies. Unless otherwise noted, all other hyperparameters are held constant across settings for fair comparison.

\noindent\textbf{Evaluation.}
% All evaluations are performed on the ImageNet validation split. For reconstruction, we report sample\mbox{-}level metrics PSNR \cite{PSNR}, SSIM \cite{SSIM}, and LPIPS \cite{LPIPS}, and distribution\mbox{-}level metrics FID \cite{FID} and sFID computed between reconstructions and the corresponding ground\mbox{-}truth images. For image generation, we report FID and IS \cite{IS} under a consistent evaluation protocol across all model variants and training schedules.
All evaluations are performed on the ImageNet validation split.
For reconstruction, we report PSNR~\cite{PSNR}, SSIM~\cite{SSIM}, and LPIPS~\cite{LPIPS} as sample-level metrics, and FID~\cite{FID} and sFID computed between reconstructions and the corresponding ground-truth images as distribution-level metrics.
For image generation, we report FID and IS~\cite{IS} under a unified evaluation protocol across all model variants and training strategies.

\begin{figure*}[t]
  \centering
  \begin{subfigure}{0.32\linewidth}
    \centering
    % \fbox{\rule{0pt}{2in} \rule{.9\linewidth}{00pt}}
    \includegraphics[width=\linewidth]{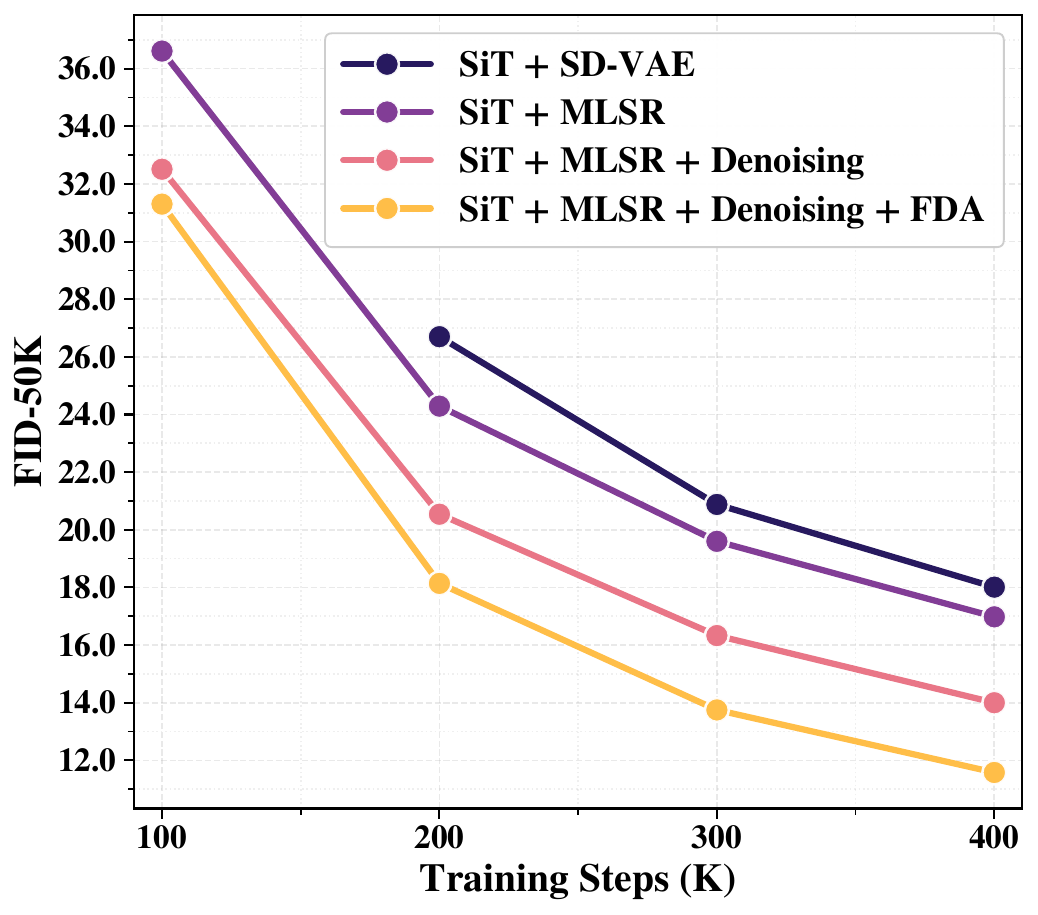}
    \caption{\textit{\textbf{ViT-based tokenizer - f16d16}}.}
    \label{fig:ab-1}
  \end{subfigure}
  \hfill
  \begin{subfigure}{0.32\linewidth}
    \centering
    % \fbox{\rule{0pt}{2in} \rule{.9\linewidth}{0pt}}
    \includegraphics[width=\linewidth]{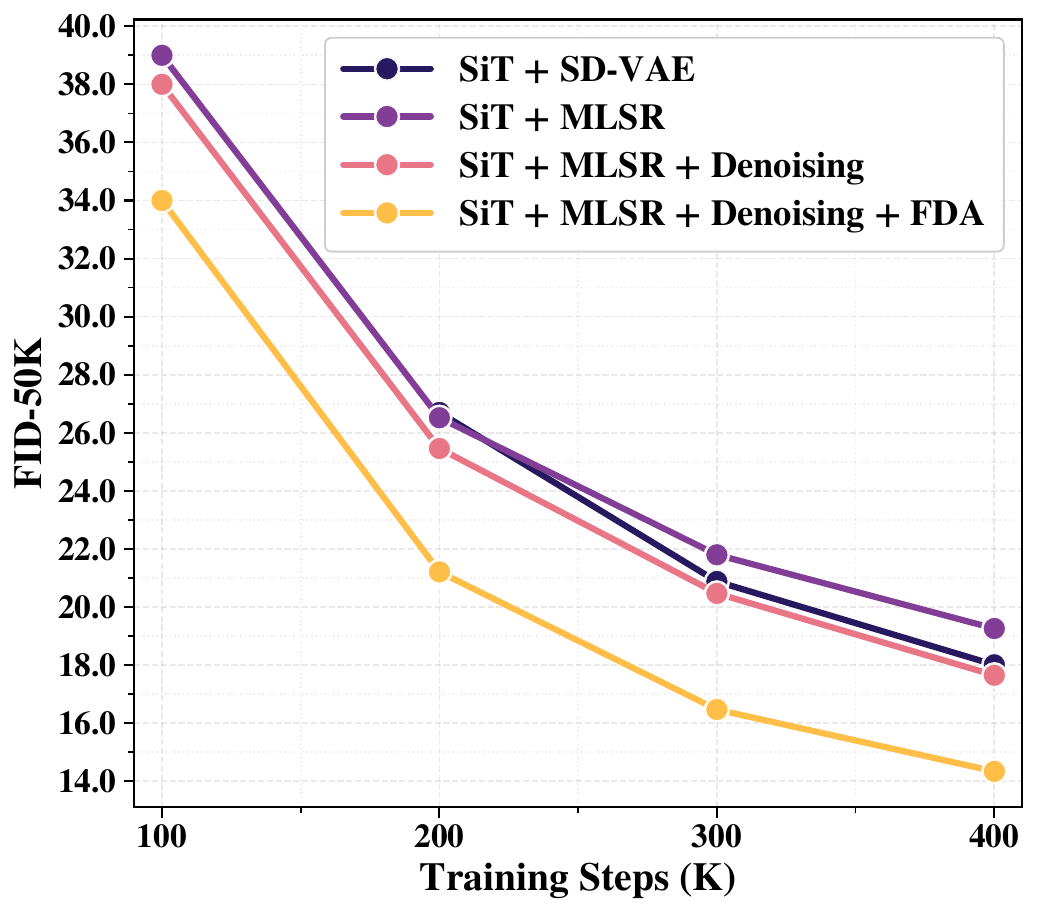}
    \caption{\textit{\textbf{ViT-based tokenizer - f16d32}}.}
    \label{fig:ab-2}
  \end{subfigure}
  \hfill
  \begin{subfigure}{0.32\linewidth}
    \centering
    % \fbox{\rule{0pt}{2in} \rule{.9\linewidth}{0pt}}
    \includegraphics[width=\linewidth]{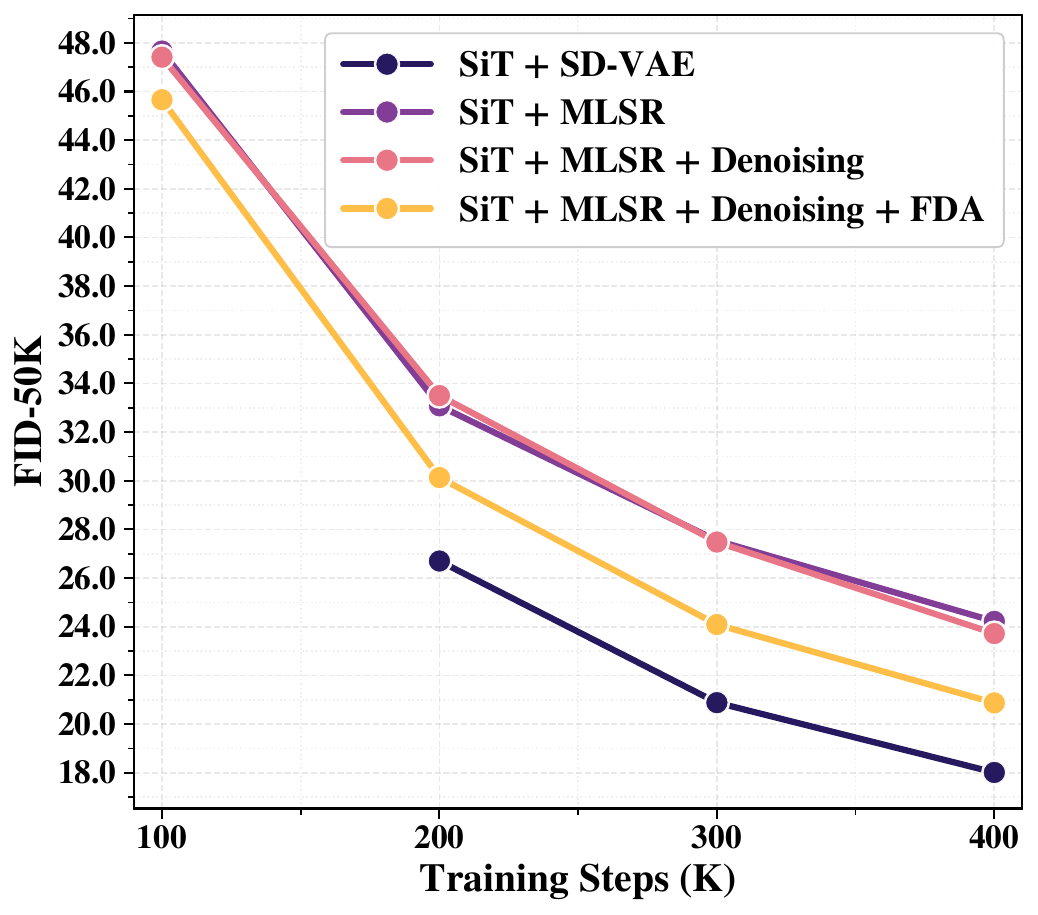}
    \caption{\textit{\textbf{ViT-based tokenizer - f16d64}}.}
    \label{fig:ab-3}
  \end{subfigure}
  \vspace{-0.5em}
  \caption{FID comparisons with vanilla SiT across different VAE settings on ImageNet $256\times256$ without CFG. 
Introducing denoising and frequency-domain alignment consistently improves convergence speed and generation quality across all latent dimensionalities.}

  \label{fig:convergence-curves}
  \vspace{-1.5em}
\end{figure*}

\subsection{Denoising Facilitates a Balance Between Reconstruction and Generation}
\label{Balance}
We conduct a comprehensive evaluation across three tokenizer configurations and nine training strategies to analyze reconstruction versus generation performance. As shown in Table~\ref{tab:regularization}, increasing latent dimensionality consistently reduces rFID but raises gFID, indicating a tradeoff where stronger tokenizers improve reconstruction at the expense of generation quality. This observation substantiates the proposed optimization dilemma.
Unless otherwise specified, "MLSR" refers to experiments conducted on improved latent representations $\mathcal{E}(x)$ obtained via spectral regularization (see Section~\ref{sec:mlsr}).
"MLSR + Denoising" denotes the use of spectrally attenuated latents $\tau \circ \mathcal{E}(x)$ to further suppress high-frequency noise (Section~\ref{sec:denoise}),
while "MLSR + Denoising + FDA" combines these denoised latents with Frequency-Domain Alignment to guide the diffusion process (Section~\ref{sec:fda}). Unless otherwise specified for ablation studies, we use the denoising level~$\ell = 2$ latent described in Section~\ref{sec:mlsr} for all experiments.

\noindent \textbf{Denoising Preserves Reconstruction Fidelity.}
% Across all latent dimensionalities, our denoised latent representations achieve reconstruction metrics that are nearly identical to those of the baseline tokenizers, with PSNR differences below \(0.05\,\mathrm{dB}\) and rFID differences under \(0.01\). This consistency indicates that denoising primarily removes high-frequency components that contribute little to perceptual quality. As shown in Figure~\ref{fig:pca-visualization}, PCA projections of the latent space exhibit smoother spatial distributions and suppressed high-frequency fluctuations, while still preserving the structural information required for accurate decoding. Together, the alignment of quantitative metrics and observed spatial smoothness suggests that denoising removes noise-like artifacts rather than meaningful content. 
% Notably, even though the denoised latent is trained with a relaxed, perceptually equivalent reconstruction target, it still achieves excellent fidelity. We hypothesize that the reduced complexity of this target yields smaller prediction errors, whereas the original latent incurs larger errors due to its high-frequency noise.
Across all latent dimensionalities, our denoised latent representations achieve reconstruction metrics nearly identical to those of the baseline tokenizers, with PSNR differences below \(0.05\,\mathrm{dB}\) and rFID differences under \(0.01\). This indicates that denoising primarily removes high-frequency components that have negligible impact on perceptual quality. As shown in Figure~\ref{fig:pca-visualization}, PCA projections of the denoised latent space exhibit smoother spatial distributions and reduced high-frequency artifacts while preserving the structural information required for accurate decoding. 
Notably, even though the denoised latent is trained with a relaxed, perceptually equivalent reconstruction target, it still achieves high fidelity. We hypothesize that this target’s reduced complexity leads to smaller prediction errors, whereas the original latent suffers larger errors due to its high-frequency noise.

\begin{figure}[t]
  \centering
    \centering
    \includegraphics[width=\linewidth]{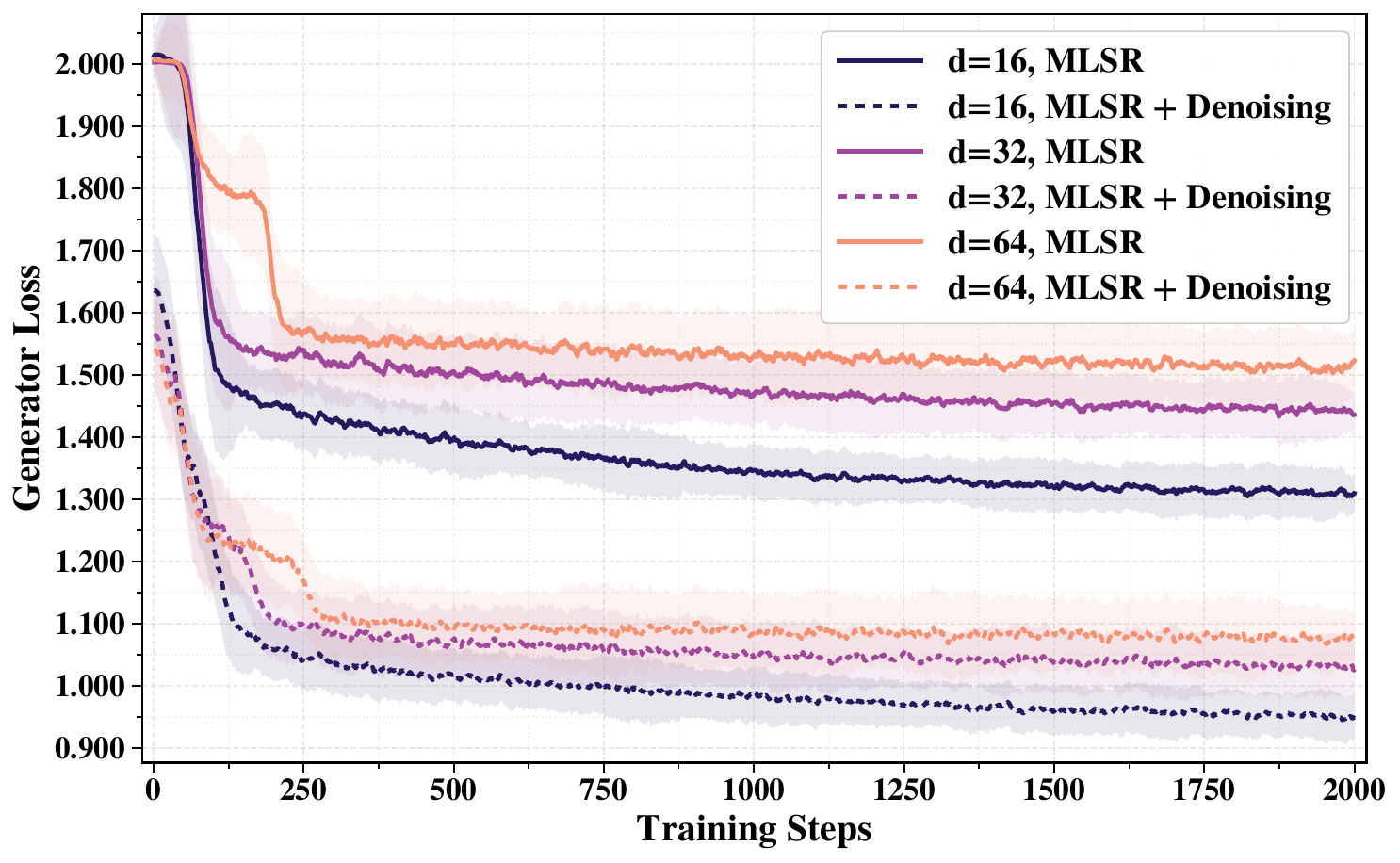}
    \vspace{-2em}
    \caption{Training losses of tokenizers under different settings. }
    \label{fig:losses}
  \hfill
  \vspace{-1.5em}
\end{figure}

%====================== Single-column table: Reconstruction only (tokens + LPIPS) ======================
\begin{table}[t]
\centering
\footnotesize
\setlength{\tabcolsep}{5.5pt}
\renewcommand{\arraystretch}{1.12}
\vspace{-0.7em}
\caption{Reconstruction performance on ImageNet $256\times256$. Metrics are computed from autoencoder outputs.}
\vspace{-0.3em}
\begin{tabular}{lccccc}
\toprule
Tokenizer & Tokens & rFID$\downarrow$ & PSNR$\uparrow$ & SSIM$\uparrow$ & LPIPS$\downarrow$ \\
\midrule
\rowcolor{red!6} \multicolumn{6}{l}{\textbf{\textit{Conv-based Tokenizer/Autoencoder}}} \\
\midrule
DC-AE \cite{DCAE}      & 64   & 0.77 & 23.93 & 0.766 & 0.092 \\
VQVAE \cite{VQVAE}      & 256  & 8.01 & 19.41 & 0.476 & 0.191 \\
MaskGIT \cite{maskgit}    & 256  & 3.79 & 18.11 & 0.427 & 0.202 \\
MaskBit \cite{maskbit}    & 256   & 1.29 & 21.07 & 0.539 & 0.142 \\
VAVAE \cite{vavae}      & 256  & 0.28 & 26.30 & 0.846 & 0.050 \\
SD-VAE \cite{LDN}     & 1024 & 0.62 & 26.04 & 0.834 & -- \\
SDXL-VAE \cite{SDXL}   & 1024 & 0.73 & 25.55 & 0.727 & -- \\
GaussToken \cite{gausstoken} & 1024 & 1.70 & 22.40 & 0.597 & 0.112 \\
EQVAE \cite{EQVAE}      & 1024 & 0.82 & 25.95    & 0.720    & 0.141 \\
\midrule
\rowcolor{red!6} \multicolumn{6}{l}{\textbf{\textit{ViT-based Tokenizer/Autoencoder}}} \\
\midrule
SoftVQ \cite{softvq}     & 64    & 0.92 & 21.93 & 0.568 & 0.115 \\
TiTok-B64 \cite{Titok}  & 64    & 1.75 & 17.01 & 0.390 & 0.263 \\
TiTok-S128 \cite{Titok} & 128   & 1.73 & 17.66 & 0.413 & 0.220 \\
MAETok \cite{maetok}     & 128   & 0.48 & 23.61 & 0.763 & 0.096 \\
One-D-Piece \cite{One-D-Piece}& 256   & 1.54 & 17.74 & 0.420 & 0.210 \\
FlexTok \cite{Flextok}    & 256   & 4.02 & 17.69 & 0.475 & 0.257 \\
TexTok \cite{Textok}     & 256   & 0.69 & 24.38 & 0.645 & -- \\
HieraTok \cite{HieraTok}   & 256   & 0.45 & --    & --    & -- \\
Ours       & 256   & \textbf{0.28} & \textbf{27.26} & 0.815 & 0.091 \\
\bottomrule
\end{tabular}
\label{tab:recon_only_single}
\vspace{-1em}
\end{table}

\noindent \textbf{Denoising Accelerates and Stabilizes Optimization.}
Beyond preserving reconstruction quality, latent denoising also brings benefits to downstream generative modeling. Across all configurations, models trained on denoised latents consistently achieve lower gFID scores under the same number of optimization steps. For example, our 32-channel VAE surpasses the original SD-VAE in generation performance. Moreover, as shown in Figure~\ref{fig:convergence-curves}, denoised latents significantly accelerate training across tokenizer variants, reducing convergence time by up to $1.6\times$ compared to unregularized baselines. These improvements suggest that suppressing high-frequency noise produces a smoother and more structured latent space, which facilitates diffusion optimization and enables faster, more stable learning. This confirms the effectiveness of spectral simplification as a regularization strategy in latent diffusion generation.

%====================== Two-column table (right-aligned numbers; no citations) ======================
\begin{table*}[t]
\centering
% \small
\footnotesize
\setlength{\tabcolsep}{5.2pt}
\renewcommand{\arraystretch}{1.12}
\caption{Comparison of reconstruction and generation performance on ImageNet at \(256\times256\) resolution. ``Params'' denotes the number of parameters in the generator (G) and tokenizer (T). Arrows indicate the direction of improvement for each metric.}

\begin{tabular}{l c c c c c c c c}
\toprule
\makecell[l]{Model} & Params (G) & Params (T) & PSNR$\uparrow$ &
rFID$\downarrow$ & gFID$\downarrow$ &
IS$\uparrow$ & Pre.$\uparrow$ & Rec.$\uparrow$ \\
\midrule
\rowcolor{blue!6}\multicolumn{9}{l}{\textit{\textbf{With semantics distillation from external pretrained vision foundation models}}} \\
\midrule
\rowcolor{red!6}\multicolumn{9}{l}{\textit{\textbf{Conv-based tokenizer}}} \\
SiT-XL + REPA \cite{REPA}          & 675M & 84M  & 26.04  & 0.62 & 1.42 & 305.7 & 0.80 & 0.64 \\
LightningDiT + VA-VAE \cite{vavae}  & 675M & 84M  & 26.30  & 0.28 & 1.35 & 295.3 & 0.79 & 0.65 \\
\rowcolor{red!6}\multicolumn{9}{l}{\textit{\textbf{ViT-based tokenizer}}} \\
SiT-XL + MAETok \cite{maetok}        & 675M & 176M & 23.61  & 0.48 & 1.67 & 311.2 &  --  & --   \\
LightningDiT + MAETok \cite{maetok}  & 675M & 176M & 23.61  & 0.48 & 1.73 & 308.4 &  --  & --   \\
\midrule
\rowcolor{blue!6}\multicolumn{9}{l}{\textit{\textbf{Without semantics distillation from external pretrained vision foundation models}}} \\
\midrule
\rowcolor{red!6}\multicolumn{9}{l}{\textit{\textbf{Conv-based tokenizer}}} \\
ADM-U \cite{ADM-U}                  & 731M &   --    &   --      &  --     & 3.94 & 186.7 & 0.82 & 0.52 \\
VDM++  \cite{VDM++}                & 2.0B   &   --    &   --      &  --     & 2.12 & 267.7 &   --   &  --    \\
% MAGVIT-v2              & 307M & 116M &        &      & 1.78 & 319.4 &      &      \\
% MDTv2-XL/2             & 676M &      &        &      & 1.58 & 314.7 & 0.79 & 0.65 \\
% CausalFusion           & 676M &      &        &      & 1.77 & 282.3 & 0.82 & 0.61 \\
% MaskGiT                & 227M & 66M  &        &      & 2.28 & 276.6 & 0.89 & 0.61 \\
SoftVQ\cite{softvq}                 & 1.4B & 72M  &   --   & 2.16 & 2.15 & 322.0 & 0.79 & 0.62 \\
VAR-d30 \cite{VAR}                & 2.0B & 109M &   --   & 0.90 & 1.92 & 323.1 & 0.82 & 0.59 \\
LlamaGen-3B \cite{llamagen}            & 3.1B & 72M  &   --   & 2.16 & 2.18 & 263.3 & 0.81 & 0.58 \\
RandAR-XXL \cite{randar}             & 1.4B & 72M  &   --   & 2.16 & 2.15 & 322.0 & 0.79 & 0.62 \\
% FlowAR-H               & 1.9B & --   &   --   & --   & 1.65 & 296.5 & 0.83 & 0.60 \\
LDM-4 \cite{LDN}                  & 400M & 55M  &   --     & 0.27 & 3.60 & 247.7 & 0.87 & 0.48 \\
MaskDiT \cite{maskdit}                & 675M & 84M  & 26.04  & 0.62 & 2.28 & 276.6 & 0.89 & 0.61 \\
DiT-XL/2 \cite{dit}               & 675M & 84M  & 26.04  & 0.62 & 2.27 & 278.2 & 0.83 & 0.57 \\
SiT-XL/2 \cite{sit}               & 675M & 84M  & 26.04  & 0.62 & 2.06 & 270.3 & 0.82 & 0.59 \\
SiT-XL + DC-AE \cite{DCAE}         & 675M & 323M & 23.85  & 0.69 & 2.84 & 311.2 & --   & --   \\
MAR-L + MAR-VAE \cite{mar}        & 479M & 66M  &   --   & 1.22 & 1.78 & 296.0 & 0.81 & 0.60 \\
MAR-H + MAR-VAE \cite{mar}        & 943M & 66M  &   --   & 1.22 & 1.55 & 303.7 & 0.81 & 0.62 \\
\rowcolor{red!6}\multicolumn{9}{l}{\textit{\textbf{ViT-based tokenizer}}} \\
MAR-L + DeTok \cite{detok}          & 479M & 171M & --     & 0.68 & 1.43 & 303.5 & 0.82 & 0.61 \\
DiT-XL + HieraTok \cite{HieraTok}      & 675M & 176M & --     & 0.45 & 1.82 &  --   & --   &  --  \\
SiT-XL + Ours          & 675M & 171M & \textbf{27.26} & \textbf{0.28} & 1.82 & 274.1  &  0.81  &  0.61    \\
% SiT-XL + Ours          & 675M & 176M & \textbf{27.26} & \textbf{0.28} & 1.82  &  273.4   &  0.81    &   0.62   \\
\bottomrule
\end{tabular}
\label{tab:main_result}
\vspace{-1em}
\end{table*}
%====================== end table ======================

\subsection{Alignment Improves Generation}
\label{sec:exp_alignment}
% As previously discussed, increasing tokenizer dimensionalities often leads to improved reconstruction but degraded generation, revealing a fundamental conflict in latent representation design. To address this challenge, we introduce Frequency\mbox{-}Domain Diffusion Alignment (FDA), which consistently improves convergence by guiding noisy latents using cleaner and spectrally simpler references.
% As shown in Table~\ref{tab:regularization} and Figure~\ref{fig:convergence-curves}, adding FDA consistently improves gFID across all tokenizer configurations and training durations, with gains remaining stable under equal training budgets. This improvement stems from a simple yet effective mechanism: spectrally simpler latents serve as guidance for their noisier counterparts, helping early denoising steps form meaningful structure more reliably. In effect, FDA introduces a progressive coarse-to-fine learning process in latent space, which accelerates convergence and improves final generative quality.

As previously discussed, increasing tokenizer dimensionalities often leads to improved reconstruction but degraded generation, revealing a fundamental conflict in latent representation design. To address this challenge, we introduce Frequency\mbox{-}Domain Diffusion Alignment (FDA), which consistently improves convergence by guiding noisy latents using cleaner and spectrally simpler references.
As shown in Table~\ref{tab:regularization} and Figure~\ref{fig:convergence-curves}, adding FDA consistently improves gFID across all tokenizer configurations and training durations, with gains remaining stable under equal training budgets. This improvement stems from a simple yet effective mechanism: spectrally simpler latents serve as guidance for their noisier counterparts, helping early denoising steps form meaningful structure more reliably. In effect, FDA introduces a progressive coarse-to-fine learning process in latent space, which accelerates convergence and improves final generative quality. In particular, FDA reduces convergence steps by up to $2\times$  compared to baselines.

\subsection{Main Results}
\begin{figure*}[t!]
  \centering
  \includegraphics[width=\linewidth]{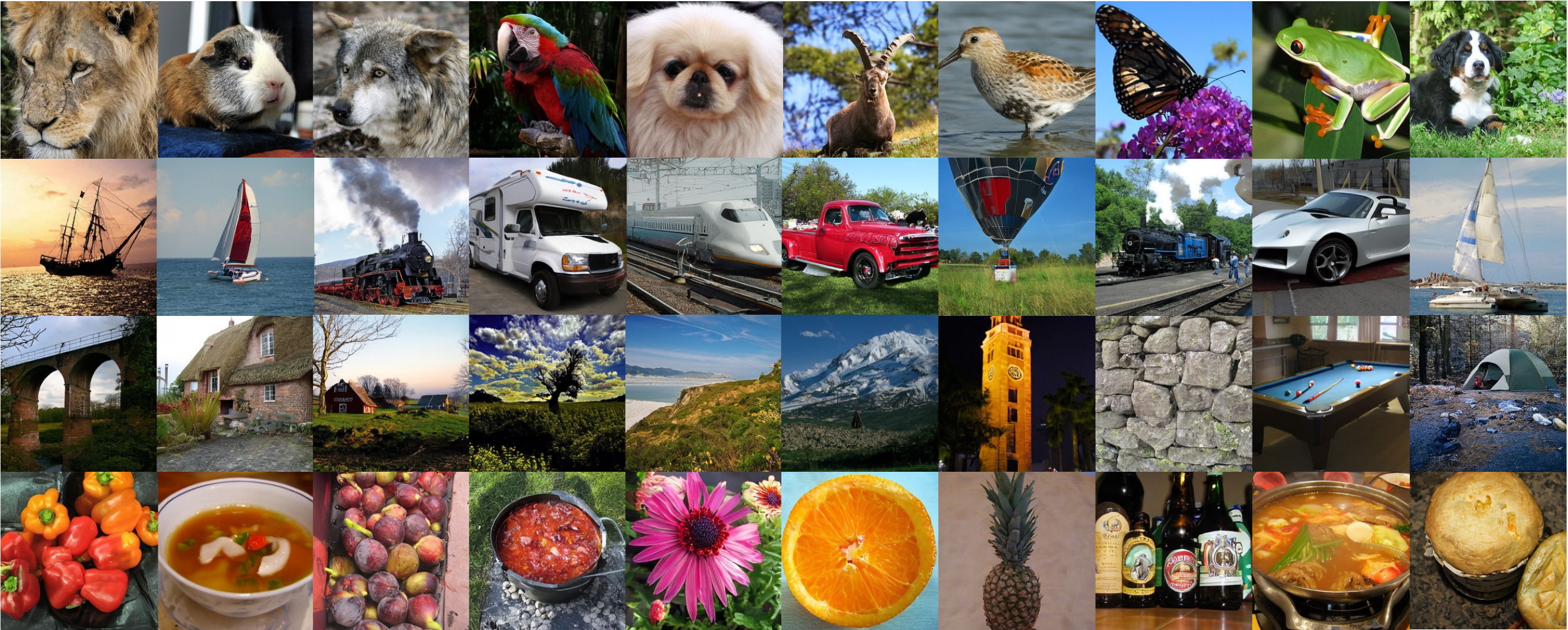}
  \caption{Visualization Results. We visualize our latent diffusion system with proposed Denoising-VAE together with SiT-XL trained on ImageNet 256 × 256 resolution using classifier-free guidance with $\omega$ = 1.8.}
    % \caption{Visualization Results. We visualize our latent diffusion system with proposed Denoising-VAE together with SiT-XL.}
  \label{fig:visimages}
  \vspace{-1.2em}
\end{figure*}

% Under a unified experimental setting on the ImageNet 256×256 benchmark, we evaluate our Denoising-VAE method against a diverse set of representative conv-based and ViT-based tokenizers.
% As shown in Table~\ref{tab:recon_only_single} and Table~\ref{tab:main_result}, our full method—comprising Denoising-VAE and Frequency-Domain Diffusion Alignment—achieves an rFID of 0.28 and a PSNR of 27.26 for reconstruction, along with a competitive gFID of 1.82 for generation. These results demonstrate that our approach closes the generation gap with strong convolutional baselines while significantly improving reconstruction quality. Compared to prior ViT-based tokenizers, our method offers superior reconstruction performance and matches or exceeds the best reported generation scores, highlighting the effectiveness of spectral denoising in preserving fidelity and enhancing compatibility with diffusion training.
% Importantly, these gains are obtained without external semantic distillation. This suggests that supervision guided by spectrally denoised latents yields a more structured and learnable target, ultimately benefiting both optimization and image quality.
Under a unified experimental setting on the ImageNet $256\times256$ benchmark~\cite{Imagenet}, we evaluate our Denoising-VAE against a diverse set of representative convolutional and ViT-based tokenizers. As evidenced by Table~\ref{tab:recon_only_single},  Table~\ref{tab:main_result} and Figure~\ref{fig:visimages}, our full method—combining Denoising-VAE with Frequency-Domain Diffusion Alignment—achieves an rFID of 0.28 and a PSNR of 27.26 for reconstruction, outperforming the strongest ViT-based alternative TexTok~\cite{Textok} by 2.88 dB in PSNR. At the same time, our method alleviates the optimization challenges inherent in high-dimensional latent spaces, substantially simplifying the training of diffusion models and significantly enhancing generative performance, achieving a competitive gFID of 1.82.
Importantly, all improvements are achieved without any external semantic distillation. This indicates that supervision derived from spectrally denoised latents provides a more structured and learnable target, ultimately benefiting both optimization dynamics and final image quality.

\subsection{Ablations and Discussions}
% \subsubsection{Denoising Levels for Training Losses}
% As shown in Figure~\ref{fig:losses}, we analyze training curves across \textbf{\textit{f16d16}}, \textbf{\textit{f16d32}}, and \textbf{\textit{f16d64}} and observe two consistent trends: first, lower-dimensional latents reach smaller losses and converge faster, indicating a reduced optimization burden in more compact latent spaces; second, at any fixed dimensionality, denoised latents maintain uniformly lower losses than their raw counterparts from the outset and throughout training. These observations jointly support our hypothesis that high-frequency suppression yields cleaner, better-conditioned latents that are easier to fit, whereas increasing dimensionality introduces additional complexity that impedes optimization. In summary, these results highlight the optimization dilemma, and show that denoising helps ease this challenge while maintaining high reconstruction fidelity and strong generative performance.

\subsubsection{Denoising Levels for Training Losses}
% As shown in Figure~\ref{fig:losses}, we analyze training curves across all latent dimensionalities, and observe two consistent trends. 
% First, lower-dimensional latents reach smaller losses and converge faster, reflecting a reduced optimization burden in more simple latent spaces. 
% Second, at fixed dimensionality, denoised latents maintain lower losses than their raw counterparts throughout training. 
% These results confirm that suppressing redundant high-frequency components produces cleaner and better-conditioned latents that are easier for diffusion models to optimize. 
% As dimensionality increases, accumulated high-frequency noise exacerbates instability and slows convergence. 
% Overall, the findings highlight an optimization dilemma in high-dimensional latent spaces and show that denoising effectively mitigates this issue while preserving reconstruction fidelity and generative quality.
As shown in Figure~\ref{fig:losses}, we analyze training curves using different latent dimensionalities VAEs and observe two consistent trends. 
First, using lower-dimensional latents to train diffusion models leads to faster convergence and lower losses, indicating that diffusion becomes substantially easier to fit in simpler latent spaces. 
Second, at a fixed dimensionality, using denoised latents to train diffusion models maintains lower losses than using raw latents.
These observations show that, as dimensionality increases, accumulated high-frequency noise amplifies instability and slows convergence. 
They further demonstrate that suppressing redundant high-frequency components produces cleaner and better-conditioned latent representations, making them easier for diffusion models to optimize.
Overall, the results highlight an inherent optimization dilemma in high-dimensional latent spaces and show that latent denoising effectively mitigates this issue while preserving both reconstruction fidelity and generative quality.

%====================== Two-column table (right-aligned numbers; no citations) ======================
\begin{table}[t]
\centering
\footnotesize
\setlength{\tabcolsep}{6pt}
\renewcommand{\arraystretch}{1.12}

\caption{Ablation study on different denoising levels. 
Moderate denoising balances reconstruction and generation.}
\vspace{-0.5em}
\begin{tabular}{lrrrr}
\toprule
Tokenizer & \multicolumn{1}{r}{rFID$\downarrow$} & \multicolumn{1}{r}{gFID$\downarrow$} & \multicolumn{1}{r}{sFID$\downarrow$} & \multicolumn{1}{r}{IS$\uparrow$} \\
\midrule
\rowcolor{red!6}\multicolumn{5}{l}{\textit{\textbf{ViT-based tokenizer - f16d16}}} \\
\midrule
MLSR                                       & 0.62 & 16.98 & 5.12 & 74.19 \\
MLSR + Denoising-level-1                  & 0.62 & 13.76 & 4.75 & 82.41 \\
MLSR + Denoising-level-2                  & 0.63 & 14.00 & 4.96 & 84.14 \\
MLSR + Denoising-level-3                  & 1.07 & 14.62 & 5.71 & 80.46 \\
\midrule
\rowcolor{red!6}\multicolumn{5}{l}{\textit{\textbf{ViT-based tokenizer - f16d32}}} \\
\midrule
MLSR                                       & 0.28 & 19.26 & 5.48 & 68.43 \\
MLSR + Denoising-level-1                  & 0.28 & 17.55 & 5.30 & 72.76 \\
MLSR + Denoising-level-2                  & 0.28 & 17.65 & 5.35 & 73.26 \\
MLSR + Denoising-level-3                  & 0.67 & 17.98 & 5.52 & 71.51 \\
\midrule
\rowcolor{red!6}\multicolumn{5}{l}{\textit{\textbf{ViT-based tokenizer - f16d64}}} \\
\midrule
MLSR                                       & 0.18 & 24.22 & 6.22 & 57.07 \\
MLSR + Denoising-level-1                  & 0.18 & 23.86 & 6.22 & 57.16 \\
MLSR + Denoising-level-2                  & 0.19 & 23.72 & 6.16 & 58.63 \\
MLSR + Denoising-level-3                  & 0.55 & 25.67 & 6.48 & 53.35 \\
\bottomrule
\label{levels}
\end{tabular}
\vspace{-3em}
\end{table}
%====================== end table =====================

\subsubsection{Denoising Levels for Generation Performance}
To systematically evaluate the effect of denoising strengths on generative performance, we conduct comparative experiments under a unified training and evaluation setup. Results in Table~\ref{levels} show that under the {\textit{f16d16}}, {\textit{f16d32}}, and {\textit{f16d64}} settings, a moderate denoising strength consistently reduces gFID, leading to smoother optimization while largely preserving reconstruction quality. When the denoising strength is further increased, gFID slightly rises in low- and mid-dimensional settings and even exceeds the no-denoising baseline in high-dimensional cases. We attribute this to the fact that excessive denoising, although making training easier and reducing loss values, shifts the latent feature distribution and weakens its structural consistency, thereby degrading both reconstruction and generation. Overall, a moderate denoising strength effectively suppresses redundant noise in high-dimensional latent spaces while preserving essential structural and energetic components, achieving a better balance between stability and representational capacity in both reconstruction and generation.

%====================== Two-column table (right-aligned numbers; no citations) ======================
\begin{table}[t]
\centering
\footnotesize
\setlength{\tabcolsep}{6pt}
\renewcommand{\arraystretch}{1.12}

\caption{Ablation study on alignment methods. 
Our FDA achieves better convergence and generation quality than other strategies.}
\vspace{-0.5em}
\begin{tabular}{lrrr}
\toprule
Tokenizer & \multicolumn{1}{r}{gFID$\downarrow$} & \multicolumn{1}{r}{sFID$\downarrow$} & \multicolumn{1}{r}{IS$\uparrow$} \\
\midrule
\rowcolor{red!6}\multicolumn{4}{l}{\textit{\textbf{ViT-based tokenizer - f16d32}}} \\
MLSR + Denoising                        & 17.65 & 5.35 & 73.26 \\
MLSR + Denoising + SRA \cite{SRA}                  & 15.44 & 5.12 & 78.90 \\
MLSR + Denoising + FDA                  & 14.34 & 5.09 & 83.28 \\
\bottomrule
\label{alignment}
\end{tabular}
\vspace{-2em}
\end{table}
%====================== end table =====================

\subsubsection{Alignment for Generation Performance}
To verify the effectiveness of our proposed spectral alignment strategy, we compare it with standard self-alignment methods. Specifically, following SRA~\cite{SRA}, we use two latents with the same denoising frequency as mutual guidance during training.  As shown in Table~\ref{alignment}, our method consistently achieves lower gFID scores than both the baseline and the SRA-based alignment method, indicating superior generation performance. We attribute this improvement to the fact that diffusion models inherently follow a coarse-to-fine generation process. Directly aligning different representations of the same latent disrupts this progression.

\subsubsection{Denoising Beats Equivariance Regularization}
To verify the advantage of the denoising approach over the equivariance-regularized method, we design and conduct a set of comparative experiments. As shown in Table~\ref{eqvs}, without applying FDA, our method significantly outperforms the equivariance-regularized baseline \cite{EQVAE} in low-dimensional latent spaces. This result indicates that, compared to indirectly constraining the representations via equivariance regularization, directly suppressing high-frequency noise in the latent space is a more effective design strategy.

%====================== Two-column table (right-aligned numbers; no citations) ======================
% \begin{table}[t]
% \centering
% \footnotesize
% \setlength{\tabcolsep}{6pt}
% \renewcommand{\arraystretch}{1.12}

% \caption{Ablation study on alignment methods. 
% Our FDA achieves better convergence and generation quality than other strategies.}

% \begin{tabular}{lrrr}
% \toprule
% Training Steps & \multicolumn{1}{r}{10k$\downarrow$} & \multicolumn{1}{r}{20k$\downarrow$} & \multicolumn{1}{r}{400k$\uparrow$} \\
% \midrule
% \rowcolor{red!6}\multicolumn{4}{l}{\textit{\textbf{Low-dimensional tokenizer}}} \\
% Spatial-Equivariance Regularized    & 15.44 & 5.12 & 78.90 \\
% MLSR + Denoising                   & 15.44 & 5.12 & 78.90 \\
% \rowcolor{red!6}\multicolumn{4}{l}{\textit{\textbf{High-dimensional tokenizer}}} \\
% MLSR + Denoising                   & 14.34 & 5.09 & 83.28 \\
% \bottomrule
% \label{eq_vs_denoise}
% \end{tabular}
% \vspace{-3em}
% \end{table}
% %====================== end table =====================

\begin{table}[t]
\centering
\footnotesize
\setlength{\tabcolsep}{5.5pt}
\renewcommand{\arraystretch}{1.12}
\caption{Comparison between EQ-VAE and Ours under a low-dimensional latent space.
Columns denote different training steps.}
\label{eqvs}
\vspace{-0.5em}
\begin{tabular}{lcccccc}
\toprule
\multirow{2}{*}{Method} 
  & \multicolumn{2}{c}{100k steps} 
  & \multicolumn{2}{c}{200k steps} 
  & \multicolumn{2}{c}{400k steps} \\
\cmidrule(lr){2-3} \cmidrule(lr){4-5} \cmidrule(lr){6-7}
& gFID$\downarrow$ & IS$\uparrow$
& gFID$\downarrow$ & IS$\uparrow$
& gFID$\downarrow$ & IS$\uparrow$ \\
\midrule
\rowcolor{red!6}\multicolumn{7}{l}{\textit{\textbf{Low-dimensional tokenizer}}} \\
EQ-VAE~\cite{EQVAE}          
  & 41.3 & 30.9 
  & 24.9 & 54.6 
  & 16.1 & 79.7 \\
Denoising-VAE             
  & 32.5 & 37.8 
  & 20.5 & 60.8 
  & 14.0 & 84.1 \\
\bottomrule
\end{tabular}
\vspace{-1.6em}
\end{table}

\vspace{-0.5em}

\section{Conclusion}
\noindent In this work, we introduce {Denoising-VAE}, a ViT-based and VFM-free autoencoder that regularizes high-dimensional latent spaces through {Spectral Self-Regularization}. Beyond the architectural efficiency gains over convolutional VAEs (reducing GFLOPs by {5.75}$\times$), we to systematically reveal that the reconstruction-generation trade-off originates from redundant high-frequency noise in high-dimensional latents. By suppressing these components, Denoising-VAE simplifies the latent space, preserves reconstruction fidelity, and substantially improves generative performance. We further propose a {Spectral Alignment} strategy that enables diffusion models trained on Denoising-VAE latents to converge nearly {2}$\times$ faster than those using SD-VAE. Extensive experiments validate the effectiveness of our framework. In future work, we plan to further investigate the intrinsic structure of high-dimensional latent spaces and develop more advanced spectral regularization techniques to better unify reconstruction and generation performance.

{
    \small
    \bibliographystyle{ieeenat_fullname}
    \bibliography{main}
}


\end{document}